\relax
\documentclass[letterpaper]{article} 
\usepackage{aaai22}  
\usepackage{times}  
\usepackage{helvet}  
\usepackage{courier}  
\usepackage[hyphens]{url}  
\usepackage{graphicx} 
\usepackage{natbib}  
\usepackage{caption} 
\DeclareCaptionStyle{ruled}{labelfont=normalfont,labelsep=colon,strut=off} 
\usepackage{algorithm}
\usepackage{algorithmic}

\usepackage{amsmath}
\usepackage{amssymb}
\usepackage{booktabs}
\usepackage{multirow}
\usepackage{pifont}

\usepackage{newfloat}
\usepackage{listings}
\floatstyle{ruled}
\newfloat{listing}{tb}{lst}{}
\floatname{listing}{Listing}
\title{CFA-Net: Controllable Face Anonymization Network with Identity Representation Manipulation}

\author{
    Tianxiang Ma \textsuperscript{\rm 1,2} \equalcontrib,
    Dongze Li\textsuperscript{\rm 1,2} \equalcontrib,
    Wei Wang \textsuperscript{\rm 1},
    Jing Dong \textsuperscript{\rm 1}
}
\affiliations{
    \{tianxiang.ma, dongze.li\}@crpiac.ia.ac.cn, \{wwang,jdong\}@nlpr.ia.ac.cn\\
    \textsuperscript{\rm 1}CRIPAC \& NLPR, CASIA\\
    \textsuperscript{\rm 2}School of Artificial Intelligence, University of Chinese Academy of Sciences\\
}
    
\usepackage{bibentry}

\begin{document}

\maketitle

\begin{abstract}
De-identification of face data has drawn increasing attention in recent years. It is important to protect people's identities meanwhile keeping the utility of the data in many computer vision tasks. We propose a Controllable Face Anonymization Network (CFA-Net), a novel approach that can anonymize the identity of given faces in images and videos, based on a generator that can disentangle face identity from other image contents. We reach the goal of controllable face anonymization through manipulating identity vectors in the generator's identity representation space. Various anonymized faces deriving from an original face can be generated through our method and maintain high similarity to the original image contents. Quantitative and qualitative results demonstrate our method's superiority over literature models on visual quality and anonymization validity.
\end{abstract}


\section{Introduction}
Rapid developments of modern biometric technologies based on deep learning bring convenience to people, while abuse of them can cause serious legal and moral issues. It has become a significant concern for people to protect their biological information, especially facial information, from being misused by unauthorized software and malicious attackers. 

Face anonymization aims to hide the identity information of a given face to protect the privacy of the corresponding person. A suitable face anonymization method should meet the following requirements. 

\begin{figure}[t]
\begin{center}
\includegraphics[width=0.9\linewidth]{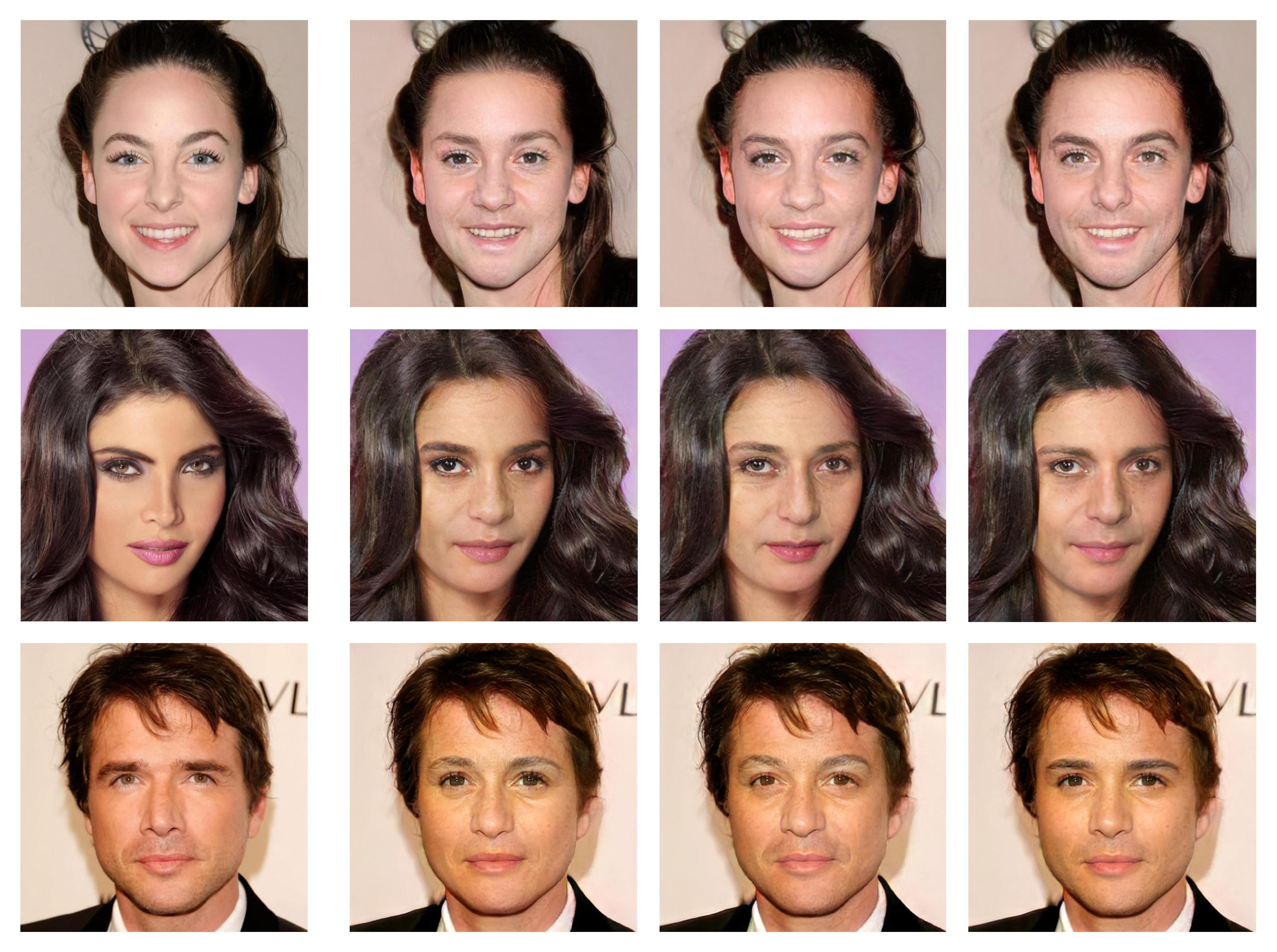}
\end{center}
    \caption{Our face anonymization results on CelebA-HQ dataset. The first column is the original images, and the remaining columns represent different anonymization effects. Identity-unrelated attributes such as pose, expression, hair cut are perfectly preserved, while identity features such as eyes, nose are modified.}
\label{fig0}
\end{figure}

\begin{itemize}
\item \textbf{Anonymous effectiveness.} A high de-identification success rate needs to be kept. An anonymized face should be able to prevent face recognition networks from matching it to the corresponding identity. 

\item \textbf{Data utility.} 
Anonymized images should look realistic to human eyes and keep their utility for downstream tasks, such as detection, tracking, and action recognition. Meanwhile, the anonymized faces should maintain high similarity with the original images and retain their attributes as much as possible, such as pose, expression, etc.

\item 
\textbf{Avoid identity leakage.} The anonymized face has to avoid the risk of revealing the identity of other people. To be more concrete, the identity distances between the anonymized face and all existing faces in the database should stay far enough so that every identity in the database is secure.

\item \textbf{Controllable.} The identity of anonymized faces needs to be precisely controlled. In other words, modifications to the original face need to be controllable to meet different demands, which is an essential point that most literature approaches have not explored. The significance of controllable anonymization is as follows. \textbf{1.} The identity distance between the anonymized face and the original face needs to be controllable so that the face recognition system can be bypassed in different threshold cases. \textbf{2.} For aesthetic reasons, users want to pick their favorite face anonymization appearance. In order to meet the above requirements, the face anonymization method needs to control the extent and variety of anonymization flexibly and accurately.


\end{itemize}
    


Traditional anonymization methods, including blurring, masking, pixelization, etc., can successfully hide the given identity, but they will seriously degrade the quality of the images and make them useless for downstream tasks. Performing direct face swapping can anonymize the target identity without damaging the image heavily, while it may cause harm to the person who provides the source identity and lead to ethical and legal issues.

In recent years, some face anonymization methods \cite{gafni2019live, maximov2020ciagan} based on generative models have gained promising results in anonymizing images and videos. However, these models have relatively weak control over anonymized faces. \cite{gafni2019live} can only yield stereotyped output for a single identity. Once the model is trained, only a fixed anonymized face can be generated for each face, and the anonymous degree cannot be adjusted. \cite{maximov2020ciagan} controls the anonymous results to some extent. However, it needs additional facial landmarks and masked images as supervisory signals, and its anonymization process depends on some reference identities of other people. In addition, the visual quality of its generated images is not good enough for human eyes. Our method explicitly separates the model training process and face anonymization into two independent stages and does not require any other reference face identity when anonymizing. Our approach can obtain an identity representation space on which we directly manipulate the identity of a given face with the maximum guarantee that the other image contents remain unchanged.



Figure \ref{fig0} shows our results on a high-resolution dataset, Celeba-HQ \cite{karras2017progressive}. With slight modifications to the original identity vector in the decoupled identity representation space, the face identity of original image is changed while identity-uncorrelated image contents are effectively preserved.

Our contributions can be summarized as follow:

\begin{itemize}

\item We propose a novel controllable face anonymization method, which can conduct flexible and precise face anonymization by directly manipulating the identity of the original face.


\item  Our approach can obtain a highly controllable identity representation space through an identity disentanglement model. With the least modification on the other image contents, various anonymized faces can be generated via manipulating original face identity vector in this space, resulting in great improvement in controllability.



\item A large number of qualitative and quantitative experiments have verified the superiority of our method against literature methods from many perspectives.

\end{itemize}

\begin{figure*}[t]
\centering
\includegraphics[width=0.9\textwidth]{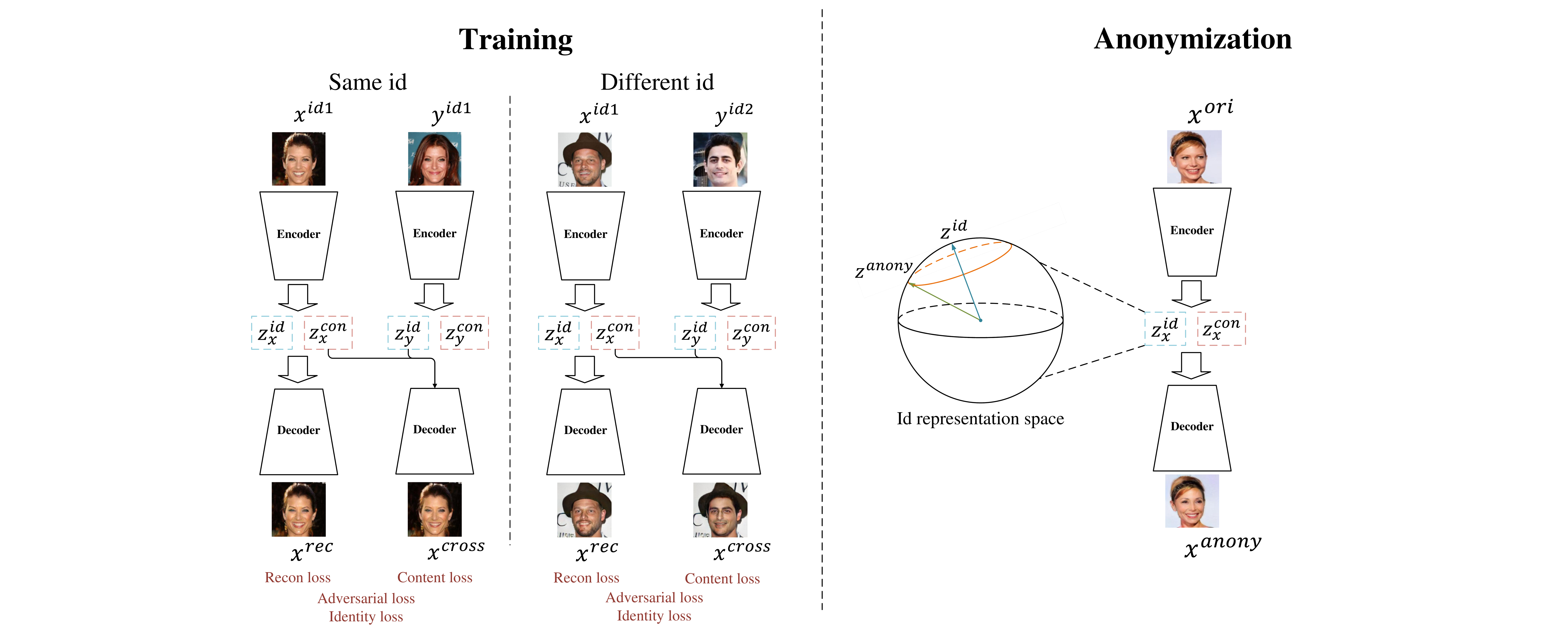}
\caption{Overview of our controllable face anonymization method. In the training phase, the model alternately samples pairs of face images of the same or different identities from the dataset and uses the corresponding loss functions to train the face identity disentanglement model. In the anonymization phase, we obtain the anonymous identity representation $\mathbf{z}^{\text{anony}}$ by sampling around the original identity vector $\mathbf{z}^{\text{id}}$ and generating anonymized face $\mathbf{x}^{\text{anony}}$.}

 

\label{fig2}
\end{figure*}

\section{Related Work}

\subsection{Face Anonymization}
Face anonymization aims to protect the private information of a given face. Simple face anonymization operations such as masking, mosaicing, blurring, etc., can destroy the usability of the face image. Nowadays, face anonymization emphasizes more on concealing the identity with least modification on other face contents. Early approaches \cite{gross2008semi, samarzija2014approach, newton2005preserving} used operations such as image distortions, image fusion, or distant face selection to achieve anonymization.  Later, \cite{jourabloo2015attribute} used an attribute classification network to extract the individual attribute parameters of the face and continuously updated them by optimization so that the generated face is not recognized as the original face. \cite{meden2017face} proposed a generative model that combines $k$ faces with different identities from the gallery data to generate anonymized faces. \cite{ren2018learning} proposed a privacy-preserving action detection model that uses a face modifier to change the identity of the input face and allows the action detection to proceed normally. \cite{sun2018natural, sun2018hybrid} adopted a two-stage training approach, with the first stage performing face identity replacement based on 3D face parameters or landmark detection and the second stage performing face image inpainting. Both methods use two-stage training and additional face annotations, which makes them less applicable. \cite{hukkelaas2019deepprivacy} proposed a generative architecture for generating anonymized faces from noise, but its generation results are not realistic enough. Recently, \cite{gafni2019live} proposed a de-identification method that can be applied to videos via an adversarial auto-encoder and a multi-level perceptual loss, but its generated anonymized faces are fixed for a given face. \cite{maximov2020ciagan} conduct diversified anonymization with a conditional GAN. However, anonymous faces generated by this method are unnatural and dissimilar from the original person. Although this method can generate multiple anonymized faces, the flexibility and continuity of control are not satisfactory. Unlike the face anonymization methods mentioned above, this paper proposes a controllable face anonymization approach, which decouples a highly controllable identity representation space. We can control the identity vectors in this space to allow flexible face anonymization while keeping other image contents unchanged maximally.




\subsection{Face Swapping}
Face swapping is also a widely concerned topic, which uses the identity of the source face to replace the target face. The widely known Deepfake is implemented using face swapping. Recently, FSGAN \cite{nirkin2019fsgan} uses several different generators to estimate the reenacted faces and their segmentation maps, and finally blends to generate swapped face images. FaceShifter \cite{li2020advancing} utilizes SPADE \cite{park2019semantic} and AdaIN \cite{huang2017arbitrary} to embed the appearance features of the target face and the identity of the source face, respectively, and uses attribute mask to fuse them to generate a realistic swapped face. Both methods directly use the identity of the source face to replace the target face, which can generate a realistic swapped face. However, direct face swapping can reveal information about the person who provides the source identity, resulting in ethical and legal issues. In contrast, our approach creates new identities that are different from all the identities in the database.


\section{Approach}

\subsection{Overview}
In this paper, we propose a controllable face anonymization network (CFA-Net), which separates the model training and anonymization process. Firstly, we train a face identity disentanglement model to decouple face identity from the identity-unrelated image contents and project it onto a unit of spherical space. Secondly, we conduct an effective face identity manipulation on the identity representation space to achieve diversified and controllable face anonymization.

\subsection{Training Identity Disentanglement model}

\subsubsection{Network Architecture}
Our identity disentanglement model is composed of a generator, two discriminators, and a pre-trained face identity extractor.

We utilize a generator that has an encoder-decoder architecture to disentangle the identity attribute of faces, as shown on the left side of figure \ref{fig2}. We separate the encoder part into two branches: the content branch and the identity branch, each of which outputs its latent code $\mathbf{z}^{\text{con}}$ and $\mathbf{z}^{\text{id}}$. They are constrained by carefully designed loss functions to force the network to learn the spatial information and the identity information, respectively. The weights of the first few layers of these two branches are shared, while the structure and weights of the last few layers are different. On the other hand, the decoder fuses the latent code of identity and content attributes to reconstruct the face image. We follow the generator structure in \cite{park2020swapping}, which uses the weight demodulation \cite{karras2020analyzing} to embed texture features into the generator. In this paper, we embed the identity representation code extracted from the encoder into the decoder in the same way.

For the identity branch of the encoder, we project the identity representation vector onto a unit of spherical space to make it easier to manipulate. To further ensure that the identity representation can contain as much identity information as possible, we utilize a pre-trained face classifier Resnet network, which is trained on the VGGFace2 \cite{cao2018vggface2} dataset, to constrain the $\mathbf{z}^{\text{id}}$ and the identity of generated faces.



Two discriminators are used to optimize the generation process. The first discriminator network $D$ serves for adversarial training and plays a minimax game with the generator. And another discriminator $D_{\text{local}}$ is employed to keep the local similarity between the generated image and the image that provides the content representation $\mathbf{z}^{\text{con}}$, which is constrained by a local patch loss. Our discriminators' architecture is the same as StyleGAN2 \cite{karras2020analyzing}, which performs well in different generative tasks.

\subsubsection{Loss Functions}

In this part, we detail the loss functions used to train our identity disentanglement model.

\noindent \textbf{Content loss.} As mentioned before, we aim to train a face identity disentanglement model that encodes the input image into identity-unrelated content features and the face identity representation vector. During training, we use a strategy of alternately introducing image pairs, where two images belong to the same person or different people, to configure more appropriate loss functions to learn the identity and content representations, as shown on the left side of Figure \ref{fig2}.

For the same identity face pair, we hope the generated face $x^{\text{cross}}$ obtained by combining $\mathbf{z}_{x}^{\text{con}}$ with $\mathbf{z}_{y}^{\text{id}}$ is still the original input face $x^{\text{id1}}$. So we directly impose the $L_{1}$ reconstruction loss as the content loss:
\begin{equation}
\mathcal{L}_{\text{\text{con}}}^{\text{same}}=\mathbb{E}_{x^{\text{\text{id1}}}, y^{\text{\text{id1}}}}\left[\left\| x^{\text{cross}} - x^{\text{id1}}\right\|_{1}\right].
\end{equation}

For the different identity face pair, The face generated by cross-combination of identity and content representations is supposed to have the identity of $y^{\text{id2}}$ and the image contents of $x^{\text{id1}}$. To constrain the contents of $x^{\text{id1}}$ and $x^{\text{cross}}$ to be similar, we introduce a local patch loss as the content loss, which can be written as:
\begin{equation}
\begin{aligned}
\min _{G} \max _{D_{\text{local}}} \mathcal{L}_{\text{\text{con}}}^{\text{diff}}=\mathbb{E}_{x^{\text{\text{id1}}},y^{\text{\text{id2}}}} &\left[-\log \left(D_{\text{local}}\left(patch\left(x^{\text{cross}}\right),\right.\right.\right.\\ &\left.\left.\left.patch\left(x^{\text{id1}}\right)\right)\right)\right],
\end{aligned}
\end{equation}
where $D_{\text{local}}$ is the discriminator used to improve the realism of local patches, and $patch$ is an operation that randomly selects image patches of different sizes from the image. The complete content constraint loss can be formulated as:
\begin{equation}
\mathcal{L}_{\text{\text{con}}} = c\cdot \mathcal{L}_{\text{\text{con}}}^{\text{same}} + \left(1 - c\right) \cdot \mathcal{L}_{\text{\text{con}}}^{\text{diff}},\ c \in \{0,1\}
\end{equation}
where $c$ is 1 for the same identity face pair and 0 for the different identity face pair.

\noindent \textbf{Identity loss.} The above loss functions keep the content similarity between the generated image $x^{\text{cross}}$ and the input image $x^{\text{id1}}$ that provides contents. Further we need to ensure the generated face $x^{\text{cross}}$ and the input image $y^{\text{id}}$ ($\text{id}\in\{\text{id1}, \text{id2}\}$) have same identity. More importantly, we need to ensure that the identity vector $\mathbf{z}^{\text{id}}$ decoupled by the encoder can express the identity information. In order to achieve the above two goals, we introduce an identity loss function, which consists of an identity consistency loss term and an identity smoothing loss term. 

The identity consistency loss aims to enhance the ability of the encoder's identity branch in learning identity information. It is achieved by matching the encoder's identity representation with the output features of a pre-trained face recognition network. This face recognition network is denoted by $R_{\text{id}}$. At the same time, it is necessary to ensure that the decoder can generate a face with the corresponding identity based on the identity representation from the encoder. The identity consistency loss can be written as:
\begin{equation}
\begin{aligned}
\mathcal{L}_{\text{id}}^{\text{consis}}{=}\mathbb{E}_{x,y}&\left[1{-}Cos\left({R}_{\text{id}}\left(x^{\text{cross}}\right), {R}_{\text{id}}\left(y^{\text{id}}\right)\right) + \right.\\&\left.1{-}Cos\left( \mathbf{z}_{y}^{\text{id}}, {R}_{\text{id}}\left(y^{\text{id}}\right)\right) \right],
\end{aligned}
\end{equation}
where $Cos(.,.)$ denotes cosine similarity between two vectors. This loss term can make the encoder's identity branch finally act like a face identity extractor and the decoder to generate an accurate face identity.

The identity smoothing loss is used to smooth the decoupled identity representation space and the identity space of the generated image. We hope that changing the identity vector $\mathbf{z}^{\text{id}}$ can result in stable variations on the generated face's identity. Perceptual path length \cite{karras2019style} can reflect the linearity and disentanglement of a model's latent space and is suitable for an identity smoothing loss term in our model. It can be expressed as:
\begin{equation}
\begin{aligned}
\mathcal{L}_{\text{id}}^{\text{smooth}}{=}\mathbb{E}_{x,y}&\left[1{-}Cos\left({R}_{\text{id}}\left(de\left(\mathbf{z}_{x}^{\text{con}},slerp\left(\mathbf{z}_{x}^{\text{id}},\mathbf{z}_{y}^{\text{id}} ; t\right)\right)\right),\right.\right. \\&
\left.\left.{R}_{\text{id}}\left(de\left(\mathbf{z}_{x}^{\text{con}},slerp\left(\mathbf{z}_{x}^{\text{id}}, \mathbf{z}_{y}^{\text{id}} ; t+\epsilon\right)\right)\right)\right)\right],
\end{aligned}
\end{equation}
where $t \sim U(0,1)$, $\epsilon=10^{-4}$, $slerp$ is the spherical interpolation operation, and $de$ represents the decoder in Figure \ref{fig2}. This loss function ensures that the identity of the generated face is smooth during manipulation on decoupled identity representation space. The complete identity loss is:
\begin{equation}
\mathcal{L}_{\text{\text{id}}} = \mathcal{L}_{\text{id}}^{\text{consis}} + \lambda \cdot \mathcal{L}_{\text{id}}^{\text{smooth}},
\end{equation}
where $\lambda$ is the hyperparameter, which we set to 10.

\noindent \textbf{Adversarial loss.} We also introduce an adversarial loss to improve the image quality through the minimax game between our generator and a discriminator. The adversarial loss term can be written as follow: 
\begin{equation}
\begin{aligned}
\min _{G} \max _{D} \mathcal{L}_{adv}=
&\mathbb{E}_{x,y}\left[\log\left(D\left(x^{\text{id1}},y^{\text{id}}\right)\right)\right]+\\
&\mathbb{E}_{x,y}\left[-\log\left(D\left(x^{\text{rec}},x^{\text{cross}}\right)\right)\right].
\end{aligned}
\end{equation}
\noindent \textbf{Reconstruction loss.} Lastly, we use a $L_{1}$ reconstruction loss between the original image and self-reconstruction output, written as:
\begin{equation}
\mathcal{L}_{\text{rec}}=\mathbb{E}_{x^{\text{\text{id1}}}}\left[\left \| x^{\text{id1}}-x^{\text{rec}}  \right \|_{1}\right].
\end{equation}

With the above all loss functions, we have well constrained the generated images' content attribute and face identity attribute. The overall training loss function of the model is:
\begin{equation}
\begin{aligned}
\mathcal{L}_{\text{overall}}=\lambda_{\text {\text{con}}}\mathcal{L}_{\text{\text{con}}}+\lambda_{\text {\text{id}}} \mathcal{L}_{\text{\text{id}}}+\lambda_{\text {adv}} \mathcal{L}_{\text{adv}}+\lambda_{\text {rec}} \mathcal{L}_{\text{rec}}
\end{aligned}
\end{equation}
where $\lambda_{\text {\text{con}}}$, $\lambda_{\text {\text{id}}}$, $\lambda_{\text {adv}}$, $\lambda_{\text {rec}}$ are the weights of the corresponding loss functions.

Through the above training scheme and loss functions, we can finally get a face identity disentanglement model, and the decoupled identity representation space has great controllability. In the following subsection, we detailed describe the manipulation method of the identity representation vector, which allows us to achieve a diverse and controllable face anonymization.

\subsection{Identity Representation Manipulation for Face Anonymization}
In this subsection, we illustrate our method of manipulating identity representation to reach the goal of controllable anonymization. Our method has a strong control on anonymous extent and can get various virtual identities. 

The anonymization process is shown on the right side of Figure \ref{fig2}. Remember, our goal is to change the original face's identity while keeping as much as possible the identity-unrelated contents. So we keep the decoupled content representation unchanged and only manipulate the identity vector. Intuitively, we search in the neighborhood of the identity vector decoupled from the original face to find a virtual identity different from the original identity but with a small amount of facial alteration. Since our identity representation space has become a unit of hypersphere due to the constraints in training, we can easily obtain various anonymous identities satisfying the requirements in the vicinity of the original identity representation following the formula below:
\begin{equation}
\begin{aligned}
\left\{
\begin{array}{l}
\left\|\mathbf{z}^{\text{anony}}\right\|_{2}=1, \vspace{2ex}\\
\frac{\displaystyle \mathbf{z}^{\text{id}}\cdot\mathbf{z}^{\text{anony}}}{\displaystyle \left\|\mathbf{z}^{\text{id}}\right\|_{2}\cdot\left\|\mathbf{z}^{\text{anony}}\right\|_{2}} < cos\left( \theta \right),
\end{array}
\right.
\end{aligned}
\label{formula1}
\end{equation}
where $\mathbf{z}^{\text{id}}$ and $\mathbf{z}^{\text{anony}}$ represent the original identity vector and the selected anonymous identity vector, respectively. $\theta$ is the preset identity threshold angle according to the face recognition network. The angle between the two identity vectors on the identity representation space is positively related to the identity distance of the two generated faces due to our model training, and its variations will bring changes to the identity of generated images. When the angle between the sampled $\mathbf{z}^{\text{anony}}$ and the original $\mathbf{z}^{\text{id}}$ is larger than a preset threshold, the identity distance between the face generated by the $\mathbf{z}^{\text{anony}}$ and the original face will be larger than the threshold of the face recognizer.  It can be seen that there exist various anonymous identity vectors, which stay in different directions but have the same angle between them and the original identity vector. This property ensures that our method can obtain diverse anonymized faces and meet users' different needs and preferences. On the other hand, our method's controllability is guaranteed by adjusting the angle between the $\mathbf{z}^{\text{anony}}$ and the $\mathbf{z}^{\text{id}}$ in decoupled identity representation space.

\section{Experiments}
\subsection{Controllable Face Anonymization}
With the proposed face identity disentanglement model and identity representation manipulation method, we can reach the goal of controllable face anonymization.

\begin{figure}[t]
\begin{center}
\includegraphics[width=1.0\linewidth]{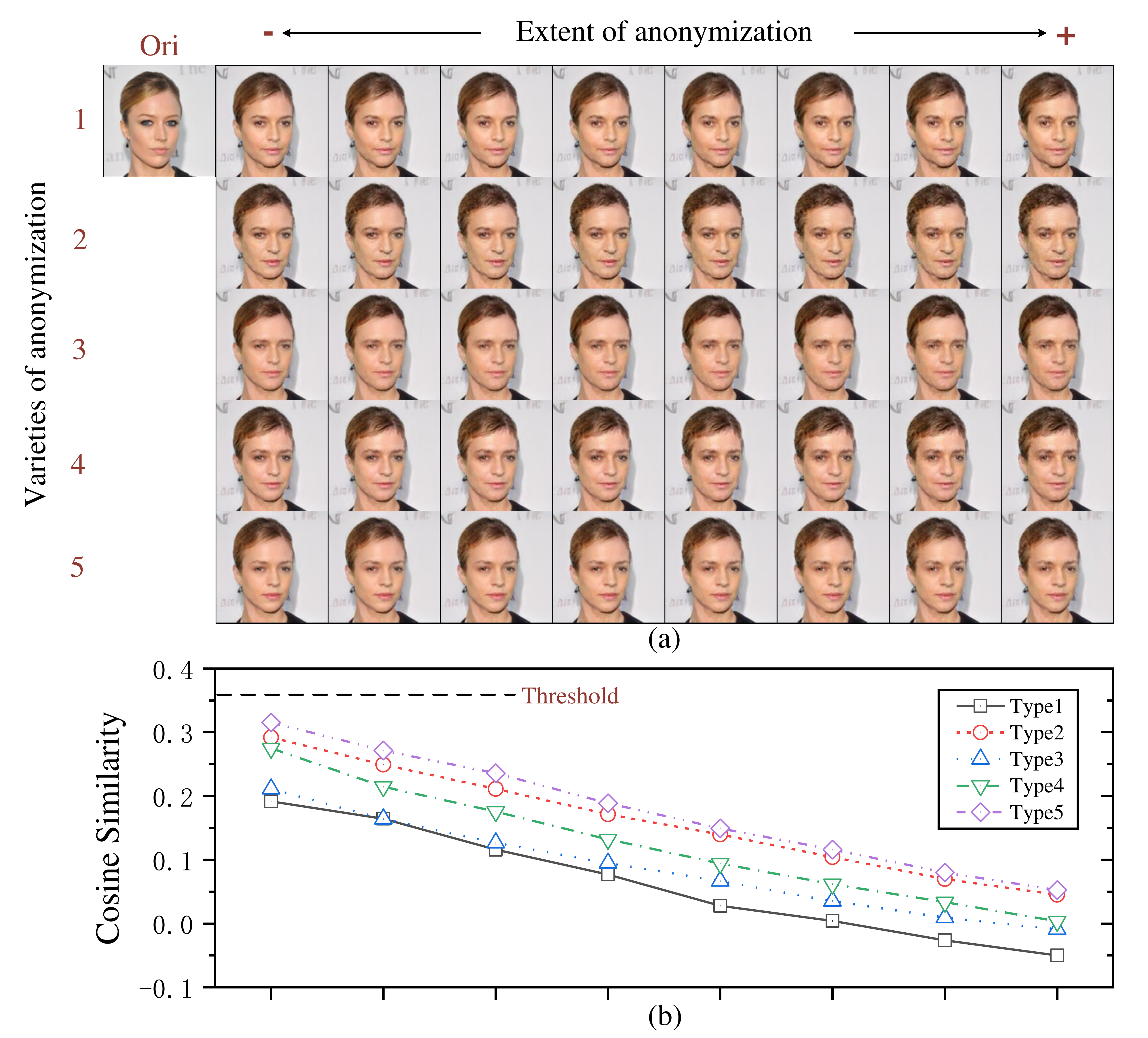}
\end{center}
\caption{Controllable face anonymization results on CelebA dataset. (a) The anonymized faces generated with $\mathbf{z}^{\text{anony}}$ have changed in two dimensions: anonymous extent and anonymous identity variety. (b) The identity distance of each generated face from the original face, which is represented by the cosine similarity of the output features in the pre-trained face recognizer. Less cosine similarity denotes a larger identity distance, and the threshold is 0.36 for face verification at the False Acceptance Rate of 0.001.}
\label{fig3}
\end{figure}

Figure \ref{fig3} shows the controllable anonymization effect of our method straightforwardly. Faces in the same row of Figure \ref{fig3}(a) is obtained by gradually increasing the angle between the original identity vector $\mathbf{z}^{\text{id}}$ and the anonymous identity vector $\mathbf{z}^{\text{anony}}$, and each row represents a different anonymous identity type. In addition, the cosine similarity between the original and the anonymized identity representation vectors in the embedding space of the pre-trained Arcface \cite{deng2019arcface} model are shown in Figure \ref{fig3}(b). It can be seen in an individual curve that the identity distance between the generated face and the original face gradually increases (cosine similarity decreasing) as the angle between $\mathbf{z}^{\text{id}}$ and $\mathbf{z}^{\text{anony}}$ increases. Cosine similarity of all points in the curve is below 0.36, the threshold of face verification, which means all the faces in the figure are successfully anonymized. On the other hand, selecting different anonymous identity representation vectors will result in different face anonymization effects as shown in each column of Figure \ref{fig3}(a).

In the following subsections, we present our experimental results and analyze key components of our model. 


\subsection{Implementation Details}
In this subsection, we present the dataset and quantitative evaluation metrics used to validate our method. The detailed training settings and network architecture can be found in our supplementary material.

\noindent \textbf{Datasets.} We use four publicly available face datasets for our experiments. \textbf{CelebA} \cite{liu2015deep}: The dataset contains 202,599 face images of 10,177 identities. We use the aligned data version, and randomly sample 8140 face identities and their corresponding images as training data. The test data are the remaining 2037 identities and their corresponding face images. We utilize this dataset to confirm the validity of our model as well as to visualize results. \textbf{LFW} \cite{huang2008labeled}: LFW is a widely used dataset for face recognition evaluation, which contains 6,000 pairs of face images, half of which are the same identities and the other half are different identities, and they are split into 10 parts for verification. We use all the same identity pairs to verify the face anonymization performance of our model and compare it with the SOTA models.
\textbf{CelebA-HQ} \cite{karras2017progressive}: We also conduct our experiments on CelebA-HQ: a dataset which is a high-quality version of CelebA that contains about 30,000 images. We use its 512$\times$512 pixel version to show our method's ability to generate diversified avatars and handling high-resolution inputs.  
\textbf{FaceForensics++} \cite{rossler2019faceforensics}: FF++ is a recently proposed forensics dataset consisting of 1000 real videos, which we use to validate that our model can easily achieve face anonymization on videos.

\noindent \textbf{Metrics.} We use true postitive rate under a fixed false accept rate to assess our anonymization strength, following test protocols in \cite{maximov2020ciagan} and \cite{gafni2019live}.
We introduced several other metrics further to show our method's superiority on image quality. Firstly, we use FID \cite{heusel2017gans}, which calculates the distribution distance between the generated images and the real images at feature level to measure the overall realism of the generate faces.
To compare the similarity between anonymized faces and real ones, we introduce LPIPS \cite{zhang2018unreasonable}, a metric that reflects human perception intuitively, to measure the similarity between two images. Traditional image similarity and quality evaluation metrics SSIM \cite{wang2004image} and PSNR are also used to measure our generated images at the pixel level.

\subsection{Face Anonymization Results}
In this subsection, we show the face anonymization effect of our approach on image and video datasets. Our method is able to achieve a high success rate of face anonymization with small changes to the original face images, and ensures that identity-unrelated face attributes remain as constant as possible. In addition, our anonymized face identity is virtual and does not match anyone's identity in the database. Additional experimental results can be found in our supplementary material.

\subsubsection{Anonymous Success Rate}
Comparison among our method, \cite{gafni2019live} and \cite{maximov2020ciagan} on anonymous success rate is shown in Table \ref{tab1}. We follow the test protocol of \cite{gafni2019live}, using all 3000 (10-fold $\times$ 300) pairs of the same identity faces in the LFW dataset and anonymizing the second face of each pair. We use pre-trained FaceNet \cite{schroff2015facenet} which shows a verification performance of over 95\% before anonymization but suffers from poor accuracy after applying all the mentioned methods, where our method surpasses others.

In addition to 1:1 verification, we further explore the face recognition ranking of several randomly chosen identities in LFW. Results are shown in Table \ref{tab2}. After anonymization, all anonymized identities are ranked outside the top 100 when matched with the original identities. What's more, the cosine similarity between anonymized faces and their best matching identities in the LFW database is smaller than the threshold, indicating that our method can create virtual identities and does not harm the identity of anyone in the database.

\begin{table}
\begin{center}
\setlength{\tabcolsep}{2mm}{
\begin{tabular}{c c c}
\toprule[1pt]
\text { Model } & \text { VGGFace2 $\downarrow$ } & \text { CASIA $\downarrow$} \\
\hline
\text { Original } & 0.986$\pm$0.010 & 0.965$\pm$0.016 \\
\text { Gafni et al. } & 0.038$\pm$0.015 & 0.038$\pm$0.011 \\
\text { Maximov et al. } & 0.034$\pm$0.016 & 0.019$\pm$0.008 \\

\text { Ours } & \textbf{0.012$\pm$0.005} & \textbf{0.010$\pm$0.007} \\
\bottomrule[1pt]

\end{tabular}}
\end{center}
   \caption{Anonymous success rate comparison with SOTA methods on LFW benchmark. The FaceNet network trained on VGGFace2 and CASIA-WebFace is used to identify the original and the anonymized faces. The True Positive Rate for a False Acceptance Rate of 0.001 is shown. }
\label{tab1}
\end{table}

\begin{table}
\begin{center}
\setlength{\tabcolsep}{1mm}{
\begin{tabular}{c c c c}
\toprule[1pt]
\text{Person} & \text{Bef-Rank} & \text{After-Rank} & \text{Cos.sim.}\\
\hline
\text { Aaron Peirsol } & 1$\pm$0 & 4619$\pm$678 & 0.33$\pm$0.01 \\
\text { Dan Kellner } & 1$\pm$0 & 3588$\pm$1385 & 0.34$\pm$0.02 \\
\text { Neri Marcore } & 1$\pm$0 & 1492$\pm$610 & 0.35$\pm$0.01 \\
\text { Ernesto Zedillo } & 1$\pm$0 & 1178$\pm$120 & 0.33$\pm$0.02 \\
\text { Flavia Pennetta } & 1$\pm$0 & 787$\pm$439 & 0.35$\pm$0.00 \\
\bottomrule[1pt]

\end{tabular}}
\end{center}
   \caption{ Ranking of face recognition at LFW database. The second column shows the ranking of each person's true identity in face recognition before anonymization, and the third column shows its ranking after anonymization by our method. The last column denotes the cosine similarity between the identity vectors of the anonymized face and its top1 matched face in the database. The threshold is 0.37 for a pre-trained Arcface at the LFW benchmark. }
\label{tab2}
\end{table}

\subsubsection{Image Quality Assessment}
Figure \ref{fig7} shows qualitative anonymization results of different methods. In figure (a), \cite{maximov2020ciagan} generates a poor quality image, which can be anonymized but cannot be adequately shared by users. \cite{gafni2019live} has less modification on faces, but it should be noticed that those three images from \cite{gafni2019live} are generated by three different models trained with different hyperparameters, while our method generates various anonymized face images with little visual artifacts by directly manipulating the decoupled original face identity vector. In figure (b), \cite{samarzija2014approach} uses a face swapping method with obvious boundary artifacts in the anonymized images, and \cite{gafni2019live} generates images with color inconsistency in the facial region. On the contrary, our approach ensures the high quality of anonymized faces and can apply in a wide range of potential application scenarios.

\begin{table*}
\begin{center}
\newcommand{\tabincell}[2]{\begin{tabular}{@{}#1@{}}#2\end{tabular}}

\setlength{\tabcolsep}{3mm}{
\begin{tabular}{c c c c c c c c}
\toprule[1pt]
\multirow{2}{*}{\text { Model }} & \multirow{2}{*}{\text { Fid $\downarrow$ }} & \multirow{2}{*}{ \text { LPIPS $\downarrow$}} & \multirow{2}{*}{ \text{SSIM $\uparrow$}} & \multirow{2}{*}{\text{PSNR $\uparrow$}} & \multicolumn{3}{c} {\text{ Detection $\uparrow$ }}\\
& & & & & \text{  MTCNN} &\text{  Dlib} & \text{SSH  } \\
\hline
Maximov et al. & 2.663 & 0.221 & 0.718 & 18.27 & 0.887 & 0.916 & \textbf{1.0} \\
\hline
\text { Ours } & \textbf{2.193} & \textbf{0.177} & \textbf{0.865} & \textbf{21.26} & \textbf{0.920} & \textbf{0.959} & \textbf{1.0} \\
\bottomrule[1pt]
\end{tabular}}
\end{center}
\caption{ Quantitative evaluation metrics. Each metric is calculated on CelebA dataset. The first four metrics are computed within the detected face region obtained by MTCNN. Three mainstream face detectors are used to evaluate the detection rate of the anonymized faces. }
\label{tab3}
\end{table*}

We further demonstrate the performance of our method through quantitative metrics, as can be seen in Table \ref{tab3}. We randomly select a total of 24,683 face images from 1200 identities in the CelebA dataset for performance metrics calculation. The first four metrics are computed within the facial region detected by the MTCNN \cite{zhang2016joint}. In addition, following \cite{maximov2020ciagan} we utilize Dlib \cite{kazemi2014one} and SSH \cite{najibi2017ssh} to evaluate the face detection rate. Our method performs not only better in FID and LPIPS that reflect image quality and human perception, respectively, but also in traditional image quality metrics, SSIM and PSNR. Images generated by our method also show a high detection rate under different face detectors, making it convenient for authorized downstream tasks such as detection and tracking.

\begin{figure}[t]
\begin{center}
\includegraphics[width=1.0\linewidth]{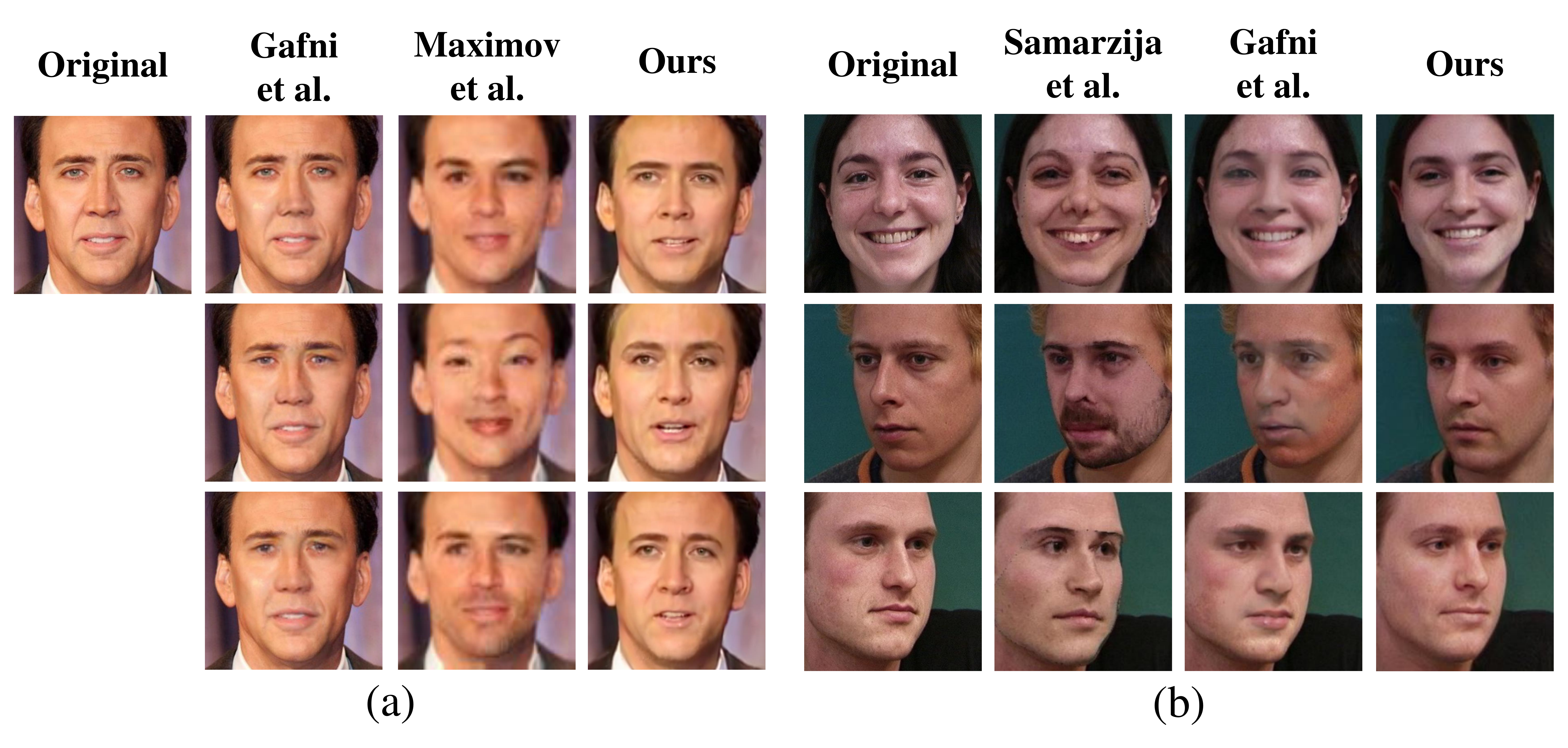}
\end{center}
    \caption{Qualitative comparison results of our method with \cite{gafni2019live}, \cite{maximov2020ciagan}, and \cite{samarzija2014approach}. }
\label{fig7}
\end{figure}

\begin{figure}[t]
\begin{center}
\includegraphics[width=0.9\linewidth]{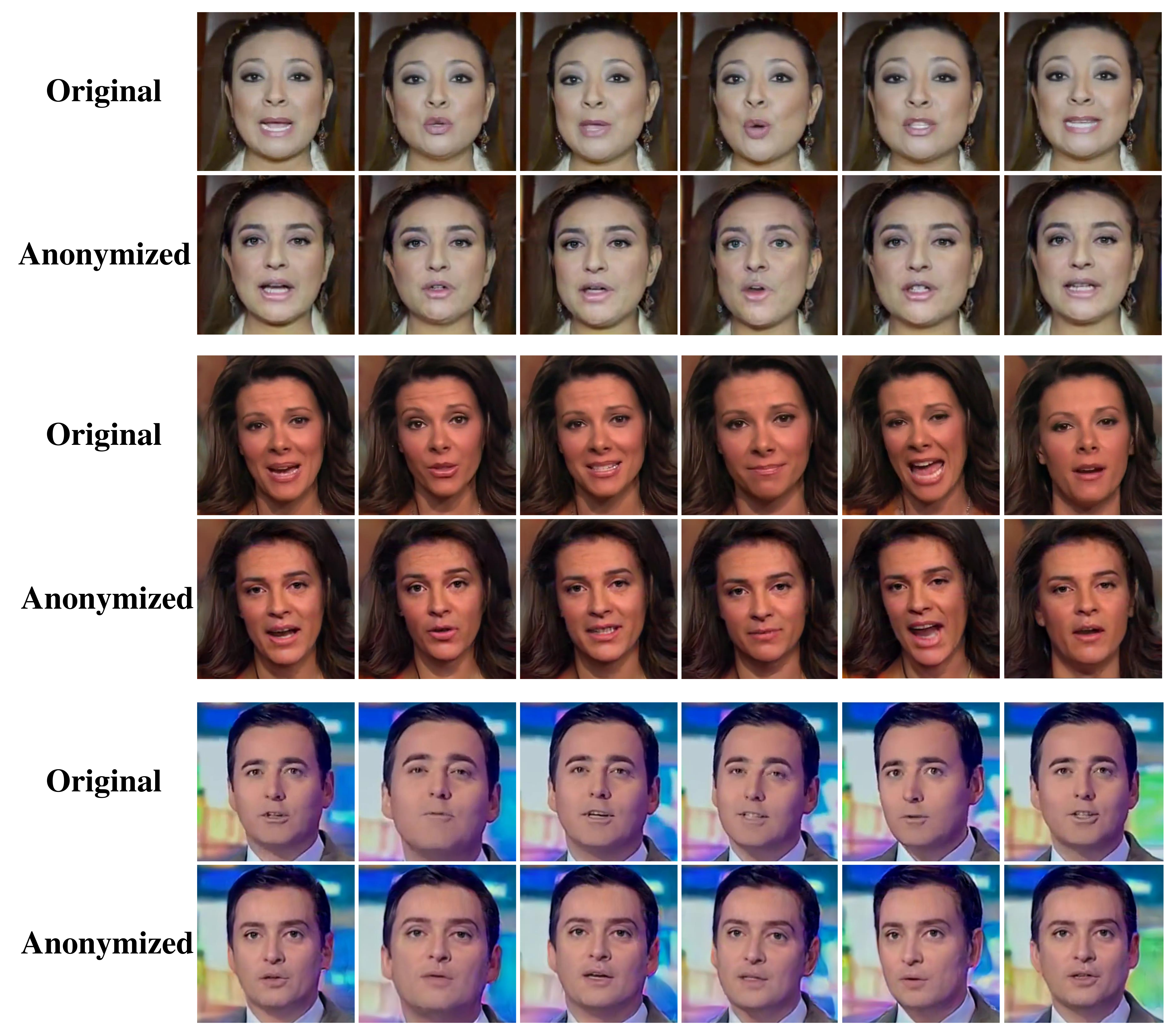}
\end{center}
    \caption{Face anonymization effects of our method on FaceForensics++ videos. The original face video frames and the corresponding anonymized faces are shown.}
\label{fig6}
\end{figure}


\begin{figure}[t]
\begin{center}
\includegraphics[width=0.7\linewidth]{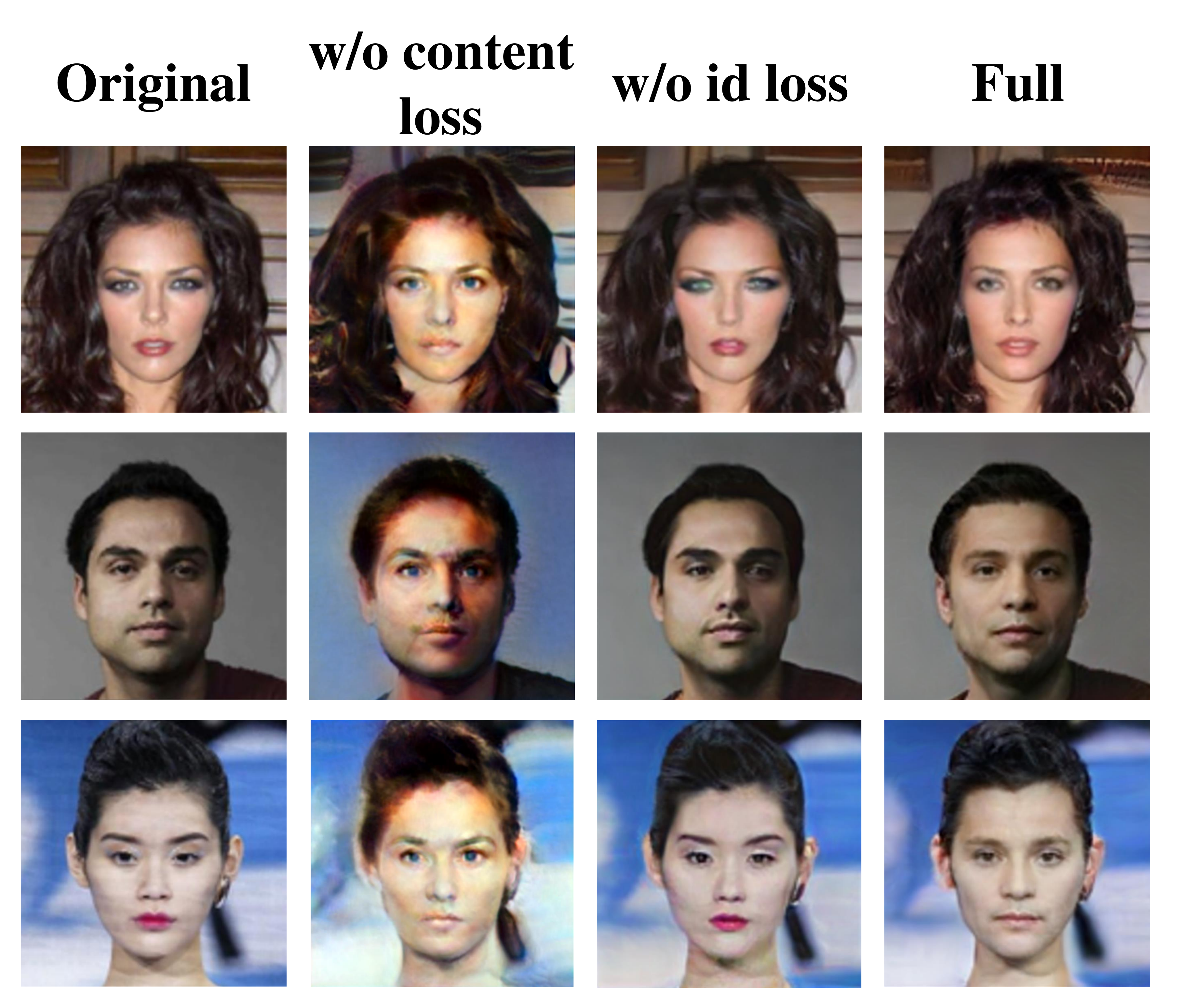}
\end{center}
   \caption{Visualization results of ablation study on key loss functions of our method.}
\label{fig10}
\end{figure}

\begin{table}[t]
\begin{center}
\setlength{\tabcolsep}{1mm}{
\begin{tabular}{c c c}
\toprule[1pt]
\text { Model } & \text { LPIPS $\downarrow$ } & \text { Anony. Success Rate $\uparrow$} \\
\hline
\text { w/o content loss } & 0.264 & 0.937 \\
\text { w/o id loss } & 0.083 & 0.004 \\

\text { Full } & 0.177 & 0.988 \\
\bottomrule[1pt]

\end{tabular}}
\end{center}
  \caption{Quantitative comparisons of ablation study on key loss functions.}
\label{tab4}
\end{table}


\subsubsection{Anonymization on Videos}
Our method can be seamlessly applied to videos. Video frames of the faces after anonymization are shown in Figure \ref{fig6}. We can see that our face anonymization method modifies the original face slightly, and the expression, pose, and other attributes of the faces change minimally. Meanwhile, the generated video frames preserve a good temporal consistency.

\subsection{Ablation Study}
In this subsection, we conduct ablation studies on the vital loss functions of our CFA-Net. Visualization of the ablation studies for different loss functions is shown in Figure \ref{fig10}. As can be seen in the figures, when content loss is missing, the generated image has the poor image quality and does not match the color and texture of the original image's content. When there is no identity loss function, the identity information cannot be extracted through its corresponding branch, making us lose control over the identity of the generated face so that the generated image is nearly the same as the original image.

The quantitative experiments in Table \ref{tab4} also indicate the importance of different loss functions. The absence of content loss causes a serious decrease in image similarity. The absence of identity loss makes the anonymization unsuccessful, and the LPIPS is quite low since the generated image is the naive reconstructed version of the original one. Best results in image quality and anonymous success rate can be reached when all the key loss functions are applied.

\section{Conclusion}
With the growing demand for data privacy protection, biological information obfuscation techniques are urgently needed, where faces have their unique status. Literature methods on face anonymization have done well in bypassing face recognition models and minimizing modifications on face images, while anonymization controlling extent and image diversity remain rarely explored, which is our proposed CFA-Net focuses on. Our method aims to obtain a highly controllable identity representation space by disentangling identity from other image contents and manipulating the decoupled identity vectors in that space to generate various anonymized faces, further achieving controllable and diversified face anonymization. Our method also surpasses previous works on anonymous success rate and image quality, which is proved by sufficient experiments.

\bibliography{aaai22}


\clearpage
\newpage
\setcounter{table}{0}
\renewcommand{\thetable}{\Alph{table}}
\setcounter{figure}{0}
\renewcommand{\thefigure}{\Alph{figure}}
\setcounter{section}{0}
\renewcommand\thesection{S\Alph{section}}

\twocolumn[
\centering
\vspace{2.0cm}
\LARGE
\textbf{Supplementary Material for CFA-Net: Controllable Face Anonymization Network with Identity Representation Manipulation} \\
\vspace{2.5cm}
]

\appendix

\section{Overview}
This supplementary material contains the following parts: network architecture, training details, and additional experimental results.

\section{Network Architecture}

\begin{figure}[h]
\setlength{\abovecaptionskip}{0.cm}
\setlength{\belowcaptionskip}{-0.cm}
\begin{center}
\includegraphics[width=0.7\linewidth]{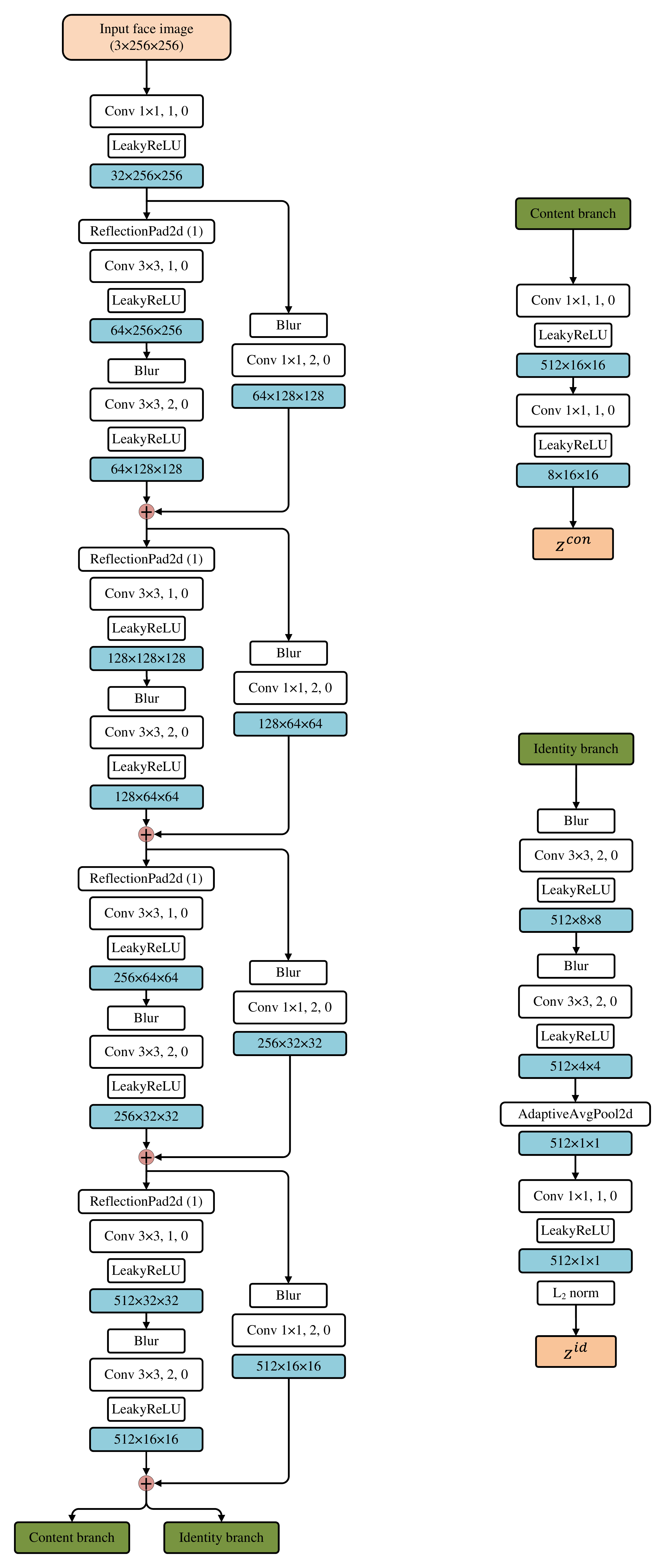}
\end{center}
    \caption{Structure of the encoder. }
\label{fig1}
\end{figure}

Our generator network is based on an encoder-decoder architecture. The network structure of the encoder is shown in Figure \ref{fig1}. The network backbone uses a residual blocks structure that leads to two branch heads, the content branch and the identity branch, which extract the identity-independent contents of the image as well as the face identity information in the image, respectively. For the decoder, we follow the generator structure in \cite{park2020swapping}, using a residual block structure in the form of convolutional layers + upsampling as the backbone, and use weight demodulation in StyleGAN2 \cite{karras2020analyzing} to embed the identity representation obtained from our encoder into the decoder to fuse the features and reconstruct the face image.

\section{Training Details}
Our approach is implemented by PyTorch using four NVIDIA TITAN RTX GPUs with 24GB memory. We adopt Adam \cite{kingma2014adam} optimizer with the learning rate of 0.002 for encoder, decoder, and discriminators. The number of training iterations is 40,000. Also, the weights of the loss functions are all set to 1, and model regularization is performed for two discriminators.
The size of the random image patch is randomly selected from 1/8 to 1/4 of the original image.
The preset identity threshold angle $\theta$ in Formula 10 needs to be set according to the threshold of the adopted face recognizer.





\section{Additional Experimental Results}

\subsection{Controllable Face Anonymization}

The additional controllable face anonymization results are shown in Figure \ref{fig3},\ref{fig4},\ref{fig5}. In these figures, (a) shows a series of anonymized faces generated by selecting different types of anonymous identity vectors $\mathbf{z}^{\text{anony}}$ and gradually increasing the angle between them and the original identity vector in our identity representation space. Curve (b) shows the cosine similarity of identity feature vectors between generated faces and the original face. We use a pre-trained face recognition network Arcface \cite{deng2019arcface} to extract the identity feature vectors from each face, and the threshold is 0.36 for face verification at the False Acceptance Rate of 0.001. From the figures, we can see that each row of faces has different identity types, reflecting the diversity of our anonymization. For each row of faces, the identity distance between the anonymized face and the original face increases with increasing the angle between $\mathbf{z}^{\text{anony}}$ and $\mathbf{z}^{\text{id}}$, and this variation is almost linear, which indicates the extent of our precise control over anonymization and the smoothness and interpretability of the identity representation space obtained by our model.

\subsection{Ablation Study}
In Figure \ref{fig6}, we explore the differences in visualization effects and anonymization controllability of different ablation models. As can be seen from the figure, the quality of the generated faces is poor when there is no content loss during training, which has some anonymity but low image availability and affects the model's control over the identity of the generated faces. When no identity loss is used during training, the generated face is simply a reconstructed image without any control over its identity, and the corresponding identity curve in (b) also shows no change in the identity of the generated faces. When using all losses training the model, our method can provide precise control over the identity of the generated images while keeping the identity-independent image contents unchanged as much as possible. Furthermore, the identity of the generated faces can be effectively manipulated by changing the $\mathbf{z}^{\text{anony}}$ in the identity representation space obtained by our method.

\subsection{Attributes Retention}
Although modifying face identity will inevitably change some face attributes, such as eyes, mouth, etc., we can still keep most of the identity-independent face attributes unchanged as much as possible since the face identity is well decoupled from the rest of the image contents. Figure \ref{fig7} shows the comparison result of the anonymized faces generated by our method and \cite{maximov2020ciagan} in terms of face attributes retention rate. From the curve, we can see that the attributes retention rate of our anonymized faces is better than \cite{maximov2020ciagan}, especially in "Big\_Lips", "Male", "Mouth\_Slightly\_Open", etc., which can reflect the decoupling ability of our method. 

\subsection{High Resolution Anonymization Results}
We also conduct experiments on high resolution datasets CelebA-HQ \cite{karras2017progressive}. Figure \ref{fig2} shows some visual anonymization results on this dataset, where the first column is the original image and the rest are the different anonymization effects obtained by our method. Each anonymization result is generated by selecting different types of anonymous identity vector $\mathbf{z}^{\text{anony}}$ in our identity representation space. The variety of anonymized faces allows users to choose different anonymized avatars according to their preferences and is also applicable to more potential application scenarios, which greatly improves the practicality of face anonymization. The high quality of the face anonymization results also demonstrates the good performance of our method.

\begin{figure}[H]
\begin{center}
\includegraphics[width=1.0\linewidth]{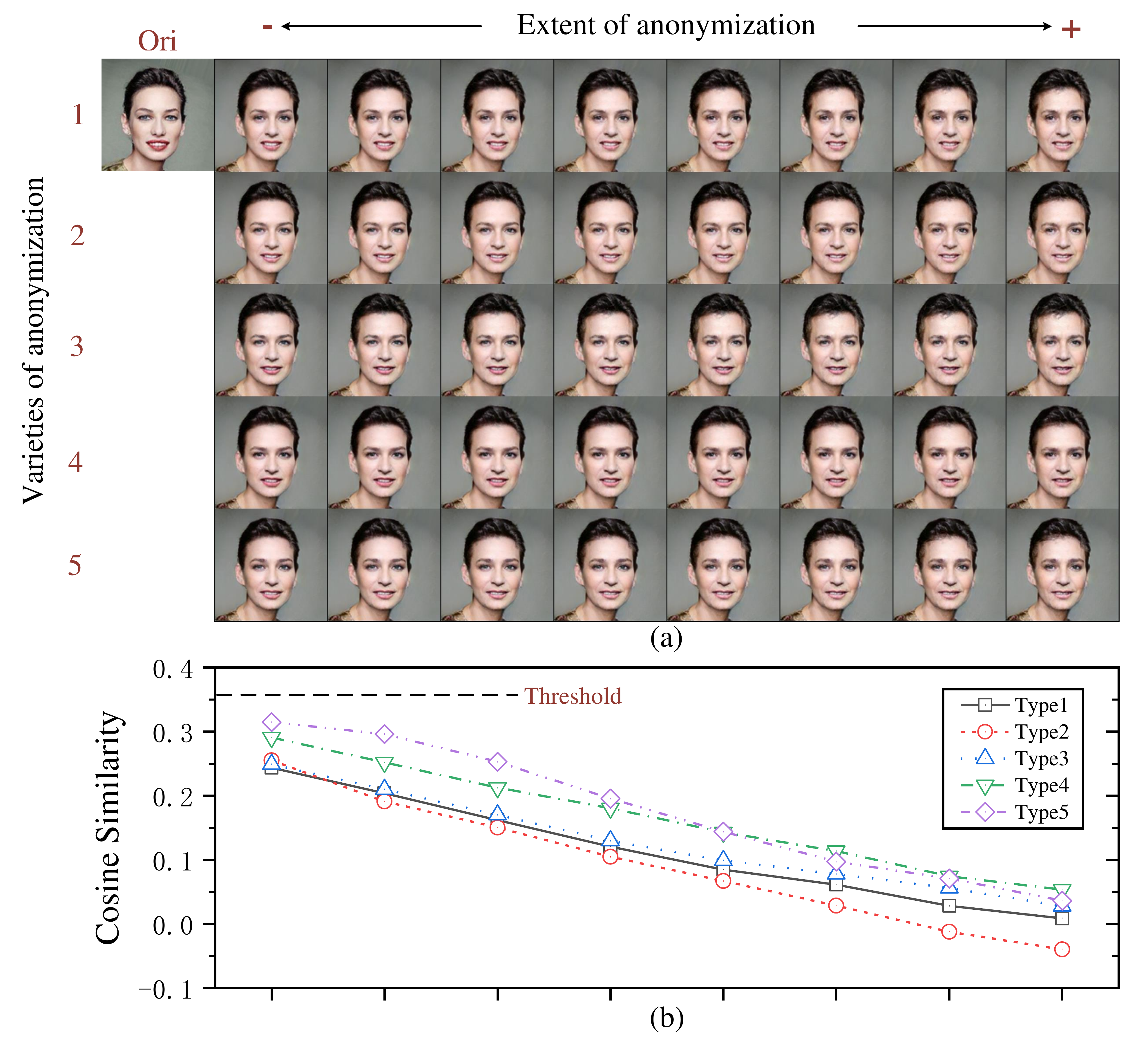}
\end{center}
    \caption{Additional controllable face anonymization results on CelebA dataset. (a) The anonymized faces generated with $\mathbf{z}^{\text{anony}}$ have changed in two dimensions: anonymous extent and anonymous identity variety. (b) The identity distance of each generated face from the original face, which is represented by cosine similarity of the output features in the pre-trained face recognizer. Less cosine similarity denotes larger identity distance, and the threshold is 0.36 for face verification at the False Acceptance Rate of 0.001. The descriptions of Figure \ref{fig4},\ref{fig5} are the same as above.}
\label{fig3}
\end{figure}

\begin{figure}[h]
\setlength{\abovecaptionskip}{0.cm}
\setlength{\belowcaptionskip}{-0.cm}
\begin{center}
\includegraphics[width=1.0\linewidth]{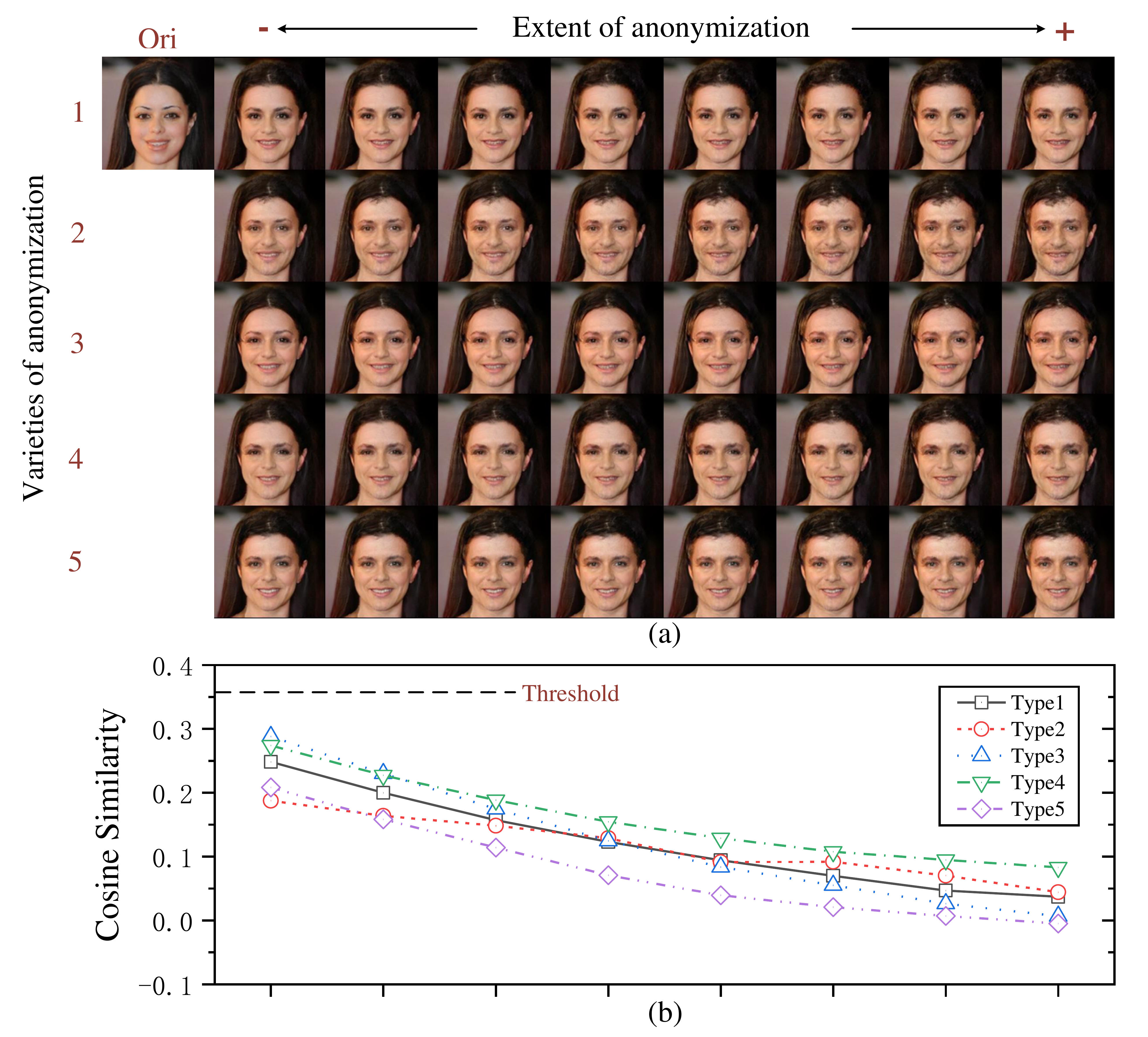}
\end{center}
    \caption{Additional controllable face anonymization results on CelebA dataset.}
\label{fig4}
\end{figure}

\begin{figure}[h]
\setlength{\abovecaptionskip}{0.cm}
\setlength{\belowcaptionskip}{-0.cm}
\begin{center}
\includegraphics[width=1.0\linewidth]{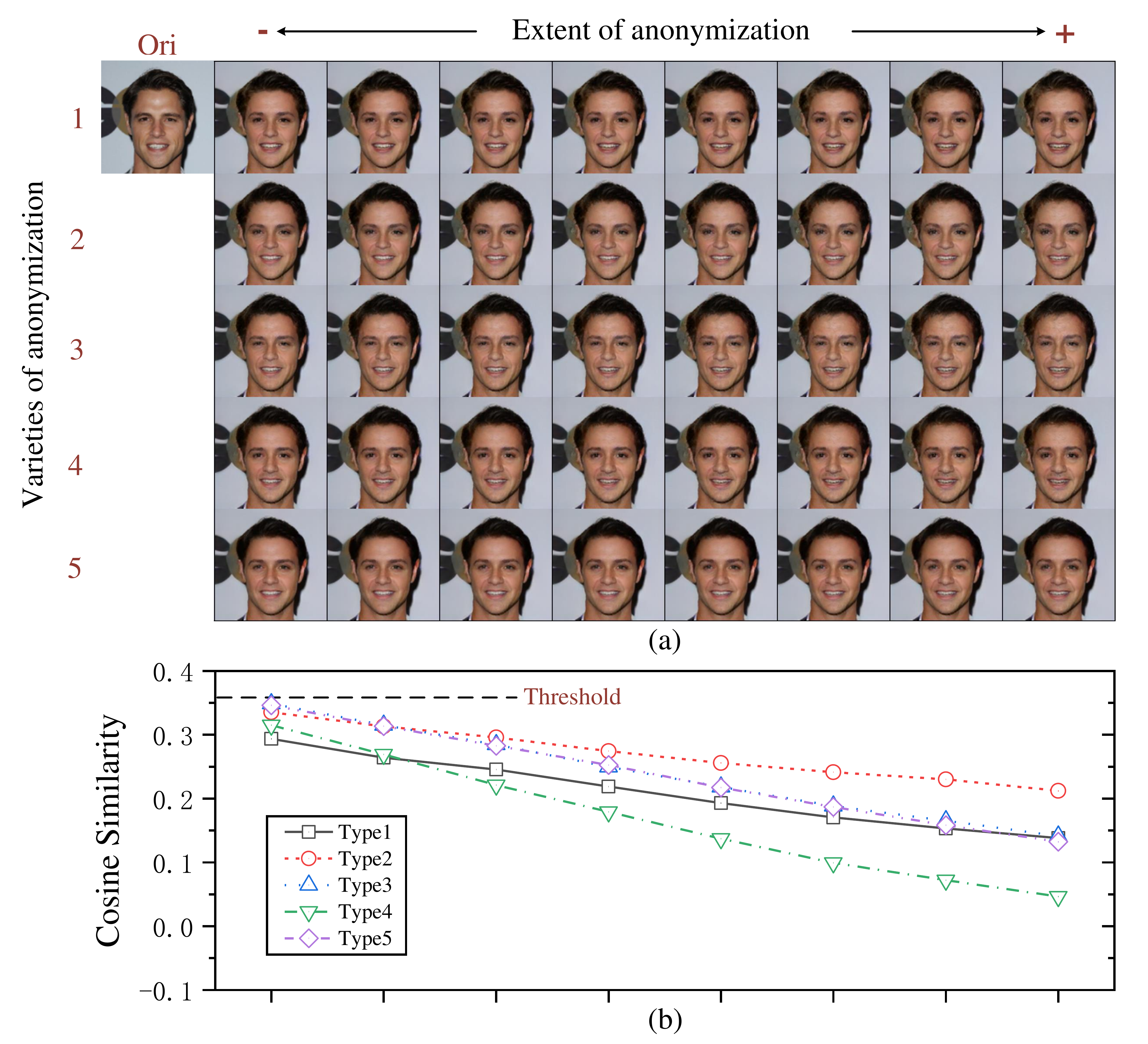}
\end{center}
    \caption{Additional controllable face anonymization results on CelebA dataset.}
\label{fig5}
\end{figure}

\begin{figure}[h]
\setlength{\abovecaptionskip}{0.cm}
\setlength{\belowcaptionskip}{-0.cm}
\begin{center}
\includegraphics[width=1.0\linewidth]{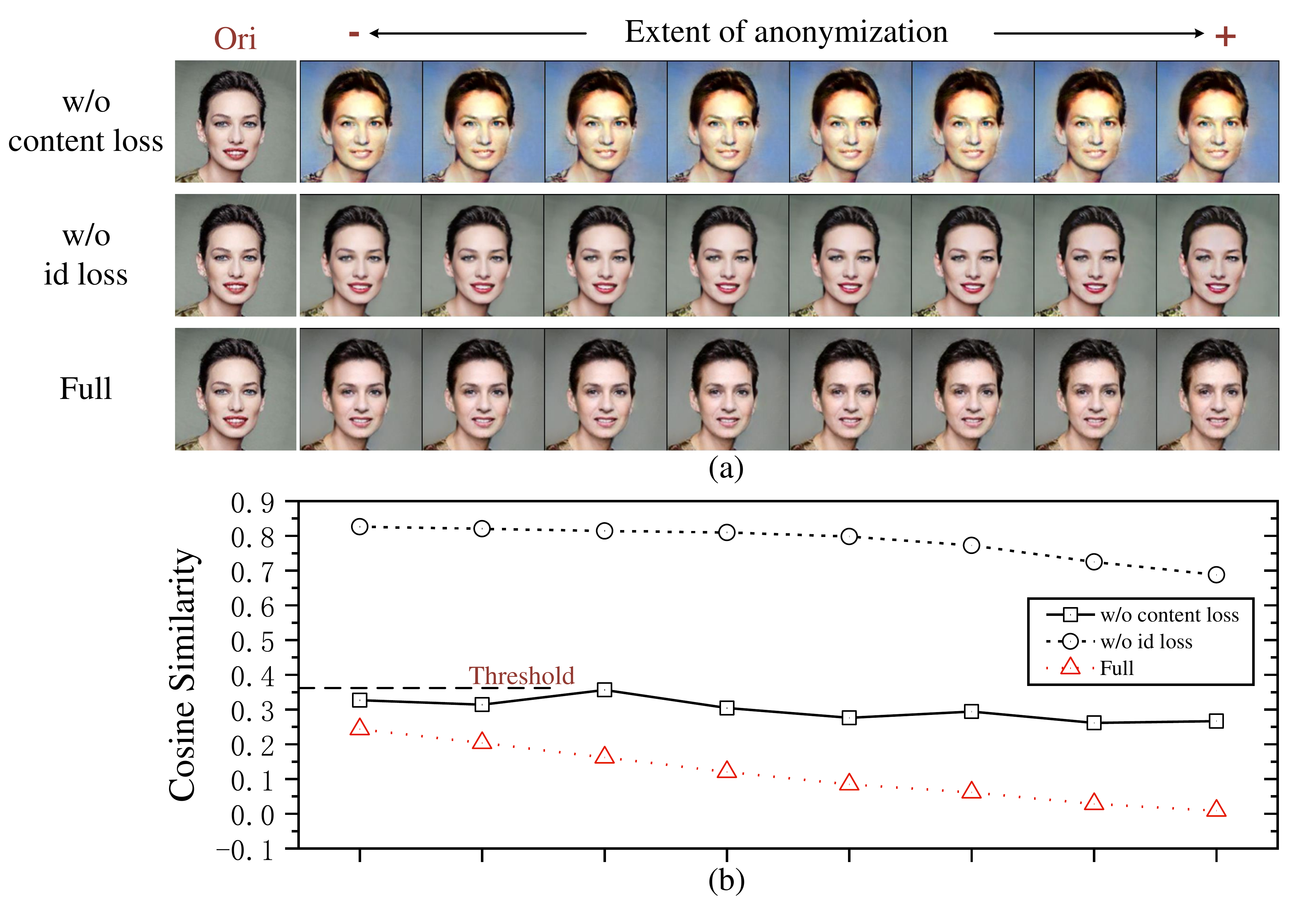}
\end{center}
    \caption{Visualization results of the ablation study and comparison of the identity controllability of the generated faces. (a) The first column are the original images, and the remaining columns are the faces generated by $\mathbf{z}^{\text{anony}}$ when gradually increasing the the angle between $\mathbf{z}^{\text{anony}}$ and $\mathbf{z}^{\text{id}}$. (b) The identity distance between each generated face and the original face, which is expressed using the cosine similarity of the identity feature vector extracted by the pre-trained face recognizer.}
\label{fig6}
\end{figure}

\begin{figure}[h]
\setlength{\abovecaptionskip}{0.cm}
\setlength{\belowcaptionskip}{-0.cm}
\begin{center}
\includegraphics[width=1.0\linewidth]{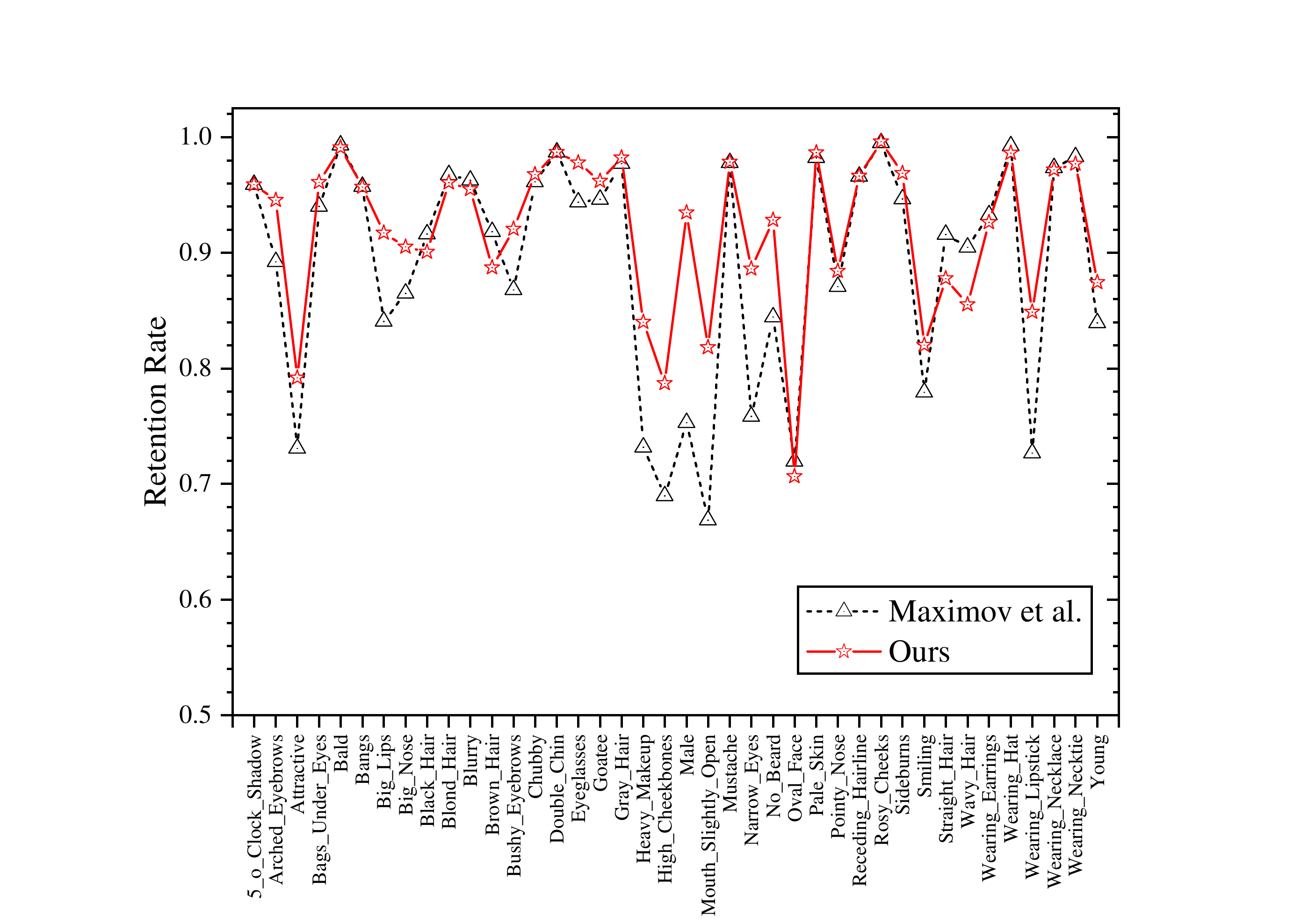}
\end{center}
    \caption{The attributes retention rate of faces after anonymization on the CelebA dataset.}
\label{fig7}
\end{figure}

\begin{figure}[H]
\setlength{\abovecaptionskip}{0.cm}
\setlength{\belowcaptionskip}{-0.cm}
\begin{center}
\includegraphics[width=1.0\linewidth]{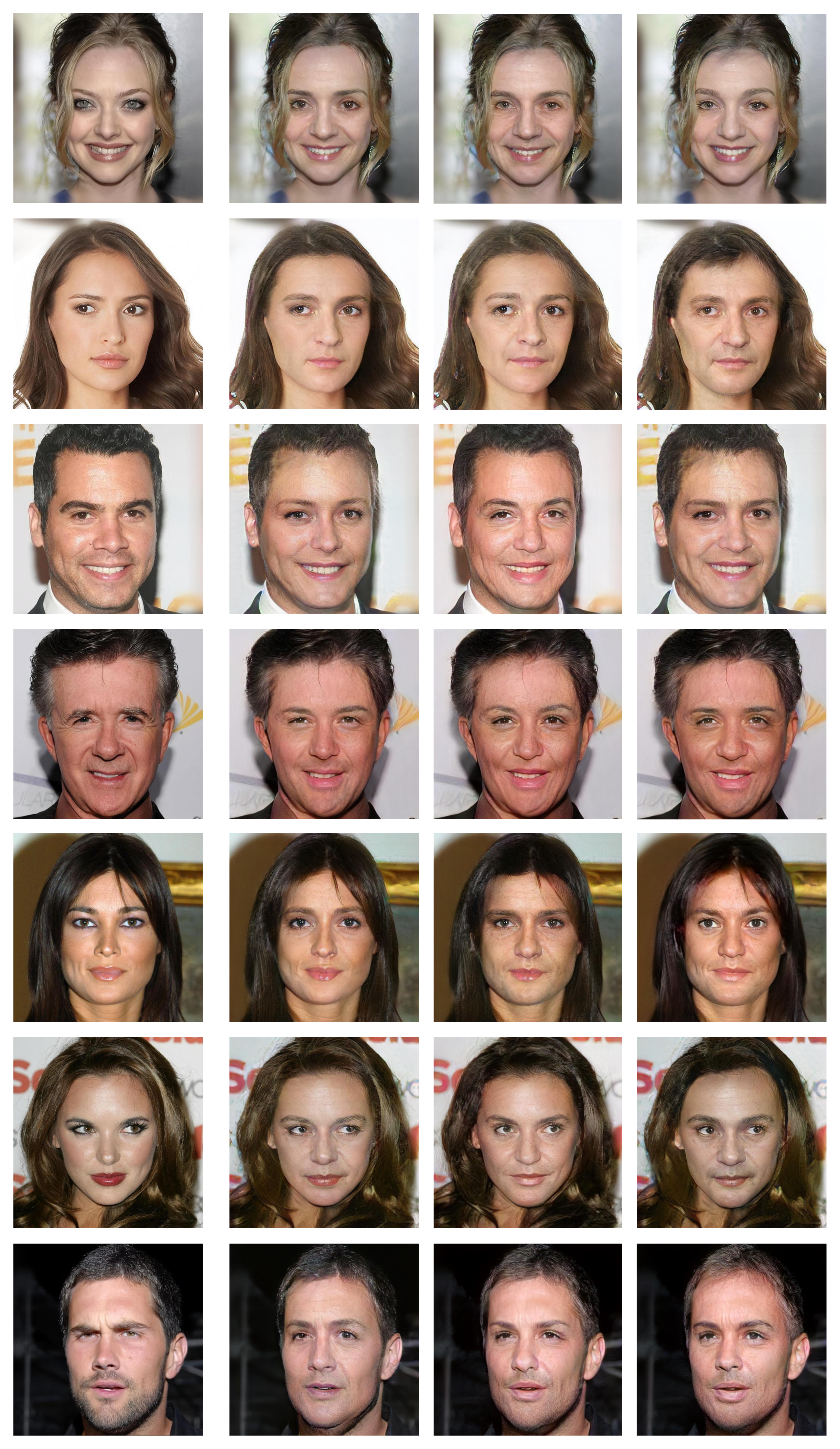}
\end{center}
    \caption{Face anonymization results of our method on the CelebA-HQ dataset. The first column is the original images, and the remaining columns represent different anonymization effects.}
\label{fig2}
\end{figure}

\end{document}